\documentclass{article}

\PassOptionsToPackage{round}{natbib}

\usepackage{enumitem}
\usepackage[preprint]{neurips_2022}

\usepackage[utf8]{inputenc} 
\usepackage[T1]{fontenc}    
\usepackage{hyperref}       
\usepackage{url}            
\usepackage{booktabs}       
\usepackage{amsfonts}       
\usepackage{nicefrac}       
\usepackage{microtype}      
\usepackage{xcolor}         
\usepackage{graphicx}
\usepackage{subfigure}
\usepackage{bm}
\usepackage{amsmath,amssymb}
\usepackage{algorithm}
\usepackage[noend]{algpseudocode}
\usepackage{soul}
\usepackage{xspace}
\usepackage{multirow}
\usepackage{makecell}

\DeclareMathOperator{\round}{round}
\DeclareMathOperator{\clamp}{clamp}

\newcommand{\cin}{C_{\text{in}}}
\newcommand{\cout}{C_{\text{out}}}

\newcommand{\SecInd}{{\sc SecInd}\xspace}
\newcommand{\SecAgg}{{\sc SecAgg}\xspace}
\newcommand{\FedAvg}{{\sc FedAvg}\xspace}
\newcommand{\SGD}{{\sc SGD}\xspace}

\def\eg{\textit{e.g.}\@\xspace}

\title{Reconciling Security and Communication Efficiency in Federated Learning}

\author{
Karthik Prasad\thanks{Equal contribution. Correspondence to \texttt{pstock@fb.com}.} $^{~\dagger}$~~ Sayan Ghosh\footnotemark[1]$^{~~\dagger}$~~ Graham Cormode$^\dagger$~~ \\ \textbf{Ilya Mironov$^\dagger$~~ Ashkan Yousefpour$^\dagger$~~ Pierre Stock$^\dagger$} \\ $^\dagger$Meta AI
}

\begin{document}

\maketitle 

\begin{abstract}

Cross-device Federated Learning is an increasingly popular machine learning setting to train a model by leveraging 
a large population of client devices with high privacy and security guarantees. 
However, 
communication efficiency remains a major bottleneck when scaling federated learning to production environments, particularly due to bandwidth constraints during uplink communication.
In this paper, we formalize and address the problem of compressing client-to-server model updates
under the Secure Aggregation primitive, a core component of Federated Learning pipelines that allows the server to aggregate the client updates without accessing them individually. 
In particular, we adapt standard scalar quantization and pruning methods
to Secure Aggregation and propose Secure Indexing, a variant of Secure Aggregation that supports quantization for extreme compression.
We establish state-of-the-art results on LEAF benchmarks in a secure Federated Learning setup with up to 40$\times$ compression in uplink communication 
with no meaningful loss in utility compared to uncompressed baselines.
\end{abstract}



\section{Introduction}
\label{sec:intro}

Federated Learning (FL) is a distributed machine learning (ML) paradigm that trains a model across a number of participating entities holding local data samples.
In this work, we focus on \emph{cross-device} FL that harnesses a large number (hundreds of millions) of edge devices with disparate characteristics such as availability, compute, memory, or connectivity
resources~\citep{kairouz2019advances}. 

Two challenges to the success of cross-device FL are privacy and scalability. 
FL was originally motivated for improving privacy since data points remain on client devices. 
However, as with other forms of ML, information about training data can be extracted via membership inference or reconstruction attacks on a trained model \citep{carlini2021membership,carlini2020extracting,watson2021importance}, or leaked through local updates~\citep{MelisSCS19,geiping2020inverting}. 
Consequently, Secure Aggregation (\SecAgg) protocols were introduced to prevent the server from directly observing individual client updates, which is a major vector for information leakage~\citep{bonavitz2019federated,huba2021papaya}. 
Additional mitigations such as  Differential Privacy (DP) may be required to offer further protection 
against attacks~\citep{dwork2006calibrating,abadi2016deep}, as discussed in Section~\ref{sec:discussion}.

Ensuring scalability to hundreds of millions of heterogeneous clients is the second challenge for FL.
Indeed, wall-clock training times are highly correlated with increasing model and batch sizes~\citep{huba2021papaya}, even with recent efforts such as FedBuff~\citep{nguyen2021federated},
and communication overhead between the server and clients dominates model convergence time.
Consequently, compression techniques were used to reduce the communication bandwidth while maintaining model accuracy.
However, a fundamental problem has been largely overlooked in the literature: in their native form, standard compression methods such as scalar quantization and pruning are not compatible with \SecAgg. 
This makes it challenging to ensure both privacy and communication efficiency.

In this paper, we address this gap by adapting compression techniques to make them compatible with \SecAgg. We focus on compressing \emph{uplink} updates from clients to the server for three reasons. 
First, uplink communication is more sensitive and so is subject to a high security bar, whereas downlink updates broadcast by the server are deemed public. 
Second, upload bandwidth is generally more restricted than download bandwidth. For instance, according to the most recent FCC report, the ratio of download to upload speeds for DSL/cable providers\footnote{FL is typically restricted to using unmetered connections, usually over Wi-Fi~\citep{huba2021papaya}.} in the US ranges between 3$\times$ to~20$\times$~\citep{fcc-broadband}.
Finally, efficient uplink communication brings several benefits beyond speeding up convergence: 
lowering communication cost reduces selection bias due to under-sampling clients with limited connectivity, improving fairness and inclusiveness. 
It also shrinks the carbon footprint of FL, the fraction of which attributable to communication can reach 95\%~\citep{qiu2021first}.

\begin{figure*}[t]
    \centering
    \includegraphics[width=\textwidth]{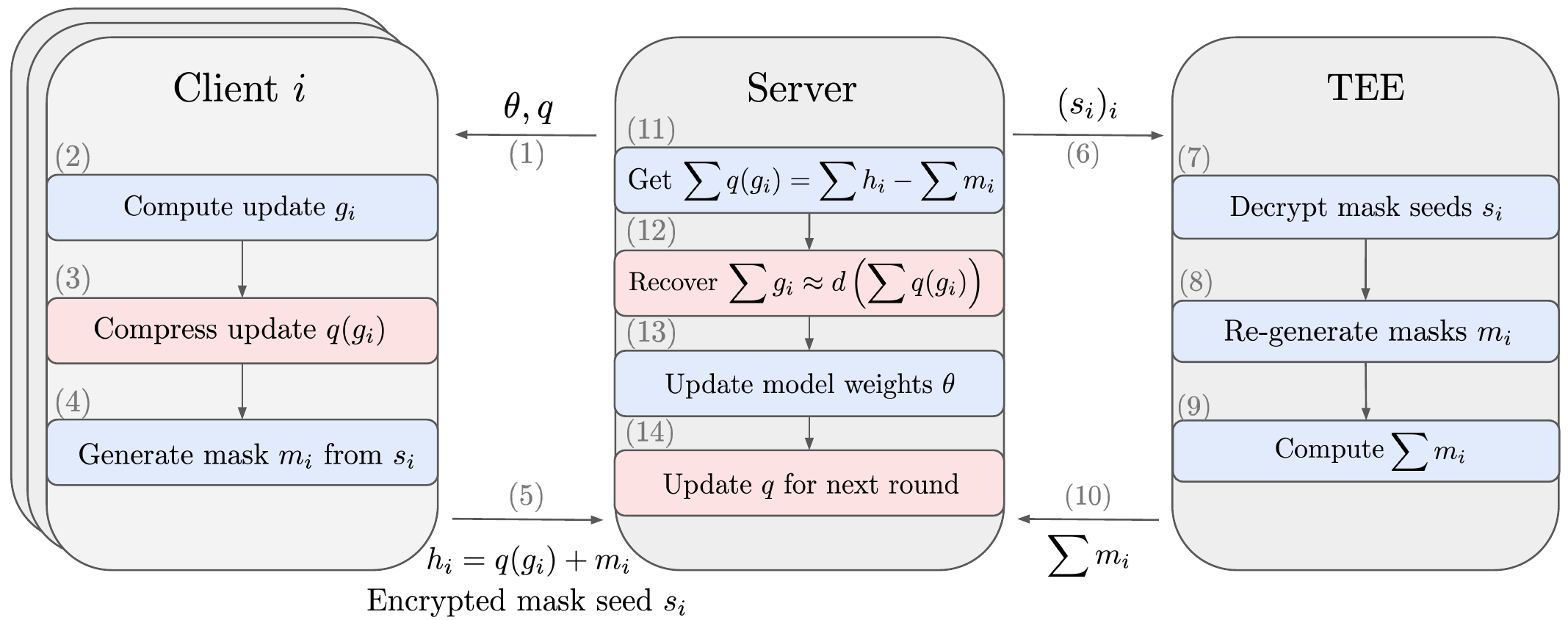}
    \vspace{-5mm}
    \caption{\label{fig:secagg_summary}
    Summary of the proposed approach for one FL round, where we omit the round dependency and DP noise for clarity. Blue boxes denote standard steps, while red boxes denote additional steps for uplink compression. Client $i$ computes local model update $g_i$, compresses it with the compression operator $q$, and encrypts it by adding a random mask $m_i$ in the compressed domain, hence reducing the uplink bandwidth (steps 2--4). The server recovers the aggregate in the compressed domain by leveraging any \SecAgg protocol (steps 7--13, illustrated here with a TEE-based \SecAgg). Since the decompression operator $d$ is linear, the server can convert the aggregate back to the non-compressed domain, up to compression error (step 12).  As with the model weights $\theta$, the parameters of the compression operator $q$ are also periodically updated and broadcast by the server (step 14). 
    In Section~\ref{sec:method}, we apply the proposed method to scalar quantization and pruning without impacting \SecAgg and propose Secure Indexing, a variant of \SecAgg for extreme uplink compression with product quantization. See Section~\ref{subsec:secagg} for details about \SecAgg and Section~\ref{sec:discussion} for a discussion on~DP.
    }
\end{figure*}

In summary, we present the following contributions in this paper: 
\begin{itemize}
    \item We highlight the fundamental mismatch between two critical components of the FL stack: \SecAgg protocols and uplink compression mechanisms.
    
    \item We formulate solutions by imposing a linearity constraint on the decompression operator, as illustrated in Figure~\ref{fig:secagg_summary} in the case of TEE-based \SecAgg.
    
    \item We adapt the popular scalar quantization and (random) pruning compression methods for compatibility with the FL stack that require no changes to the \SecAgg protocol.
    
    \item For extreme uplink compression without compromising security, we propose Secure Indexing (\SecInd), a variant of \SecAgg that supports product quantization. 
\end{itemize}

\newcommand{\para}[1]{\noindent \textbf{#1}}

\section{Related Work}
\label{sec:related}

Communication is identified as a primary efficiency bottleneck in FL, especially in the cross-device FL setting \citep{kairouz2019advances}. This has led to significant interest in reducing FL's communication requirements. In what follows, we might refer to any local model update in a distributed training procedure as a \emph{gradient}, including model updates computed following multiple local training steps.

\para{Efficient Distributed Optimization.} There is a large body of literature on reducing the communication cost for distributed training. \cite{seide2014bit} proposes quantizing gradients to one bit while carrying the quantization error forward across mini-batches with error feedback. Similarly, \cite{wen2017terngrad} proposes layer-wise ternary gradients and \cite{bernstein2018signsgd} suggests using only the sign of the gradients. Gradient sparsity is another related area that is extensively studied \citep{wangni2017gradient,aji2017sparse,lin2017deep,renggli2018sparcml,parcollet2022zerofl}. 
For instance, \cite{chen2017adacomp} and \cite{han2020adaptive} explore adapting the degree of sparsity to the distribution of local client data. Another method, QSGD, tunes the quantization level to trade possibly higher variance gradients for reduced communication bandwidth~\citep{alistarh2016qsgd}. Researchers also studied structured and sketched model updates \citep{konen2016federated}. 
For example, \cite{wang2018atomo} proposes expressing gradients as a linear combination of basis vectors common to all workers and \cite{wang2022fedlite} propose to cluster the gradients and to implement error correction on the client side. Besides gradient compression, other methods such as~\cite{vepakomma2018split,hu2019dynamic} proposes reducing the communication cost by partitioning the model such that each client learns a portion of it, while \cite{he2020group} proposes training small models and periodically distilling them to a larger central model. However, as detailed in Section~\ref{sec:background} and below, most of the proposed methods are not readily compatible with \SecAgg and cannot be used in secure FL.

\para{Bi-directional Compression.} In addition to uplink gradient compression, a line of work also focuses on downlink model compression. In a non-distributed setup, \cite{zhou2016dorefanet, courbariaux2015binaryconnect} demonstrates that it is possible to meaningfully train with low bit-width models and gradients. In FL, \cite{jiang2019model} proposes adapting the model size to the device to reduce both communication and computation overhead. Since the local models are perturbed due to compression, researchers propose adapting the optimization algorithm for better convergence \citep{liu2019double,sattler2019robust,tang2019doublesqueeze,zheng2019communicationefficient,amiri2020federated,philippenko2021preserved}.
Finally, pre-conditioning models during FL training can allow for quantized on-device inference, as demonstrated for non-distributed training by \cite{gupta2015deep, krishnamoorthi2018quantizing}. As stated in Section~\ref{sec:intro}, we do not focus on downlink model compression since uplink bandwidth is the main communication bottleneck and since \SecAgg only involves uplink communication.

\para{Aggregation in the Compressed Domain.} In the distributed setting, \cite{yu2018gradiveq} propose to leverage both gradient compression and parallel aggregation by performing the \emph{ring all-reduce} operation in the compressed domain and decompressing the aggregate. To do so, the authors exploit temporal correlations of the gradients to design a linear compression operator. 
Another method, PowerSGD~\citep{vogels2019powersgd}, leverages a fast low-rank gradient compressor. However, both aforementioned methods are not evaluated in the FL setup and do not mention \SecAgg. Indeed, the proposed methods focus on decentralized communication between the workers by leveraging the all-reduce operation. Moreover, Power SGD incorporates (stateful) error feedback on all distributed nodes, which is not readily adaptable to cross-device FL in which clients generally participate in a few (not necessarily consecutive) rounds.
Finally, \cite{rothchild2020fetchsgd} proposes FetchSGD, a compression method relying on a CountSketch, which is compatible with \SecAgg.


\newcommand{\parai}[1]{\noindent\textit{#1}}

\section{Background}
\label{sec:background}

In this section, we recall the \SecAgg protocol first, then the compression methods that we wish to adapt to \SecAgg, namely, scalar quantization, pruning, and product quantization.

\subsection{Secure Aggregation}
\label{subsec:secagg}

\SecAgg refers to a class of protocols that allow the server to aggregate client updates without accessing them individually. While \SecAgg alone does not entirely prevent client data leakage, it is a powerful and widely-used component of current at-scale cross-device FL implementations~\citep{kairouz2019advances}. Two main approaches exist in practice: software-based protocols relying on Multiparty Computation (MPC)~\citep{bonavitz2019federated,bell2020secure,LightSecAgg}, and those that leverage hardware implementations of Trusted Execution Environments (TEEs)~\citep{huba2021papaya}. 

\SecAgg relies on additive masking, where clients protect their model updates $g_i$ by adding a uniform random mask $m_i$ to it, guaranteeing that each client’s masked update is statistically indistinguishable from any other value. 
At aggregation time, the protocol ensures that all the masks are canceled out. For instance, in an MPC-based \SecAgg, the pairwise masks cancel out within the aggregation itself, since for every pair of users $i$ and $j$, after they agree on a matched pair of input perturbations, the masks $m_{i,j}$ and $m_{j,i}$ are constructed so that $m_{i,j}=-m_{j,i}$.
Similarly and as illustrated in Fig.~\ref{fig:secagg_summary}, in a TEE-based \SecAgg, the server receives $h_i = g_i + m_i$ from each client as well as the sum of the masks $\sum_i m_i$ from the TEE and recovers the sum of the updates as
\begin{equation*}
      \sum_i g_i = \sum_i h_i - \sum_i m_i.
\end{equation*}
We defer the discussion of DP noise addition by \SecAgg protocols to Section~\ref{sec:discussion}.

\para{Finite Group.}
\SecAgg requires that the plaintexts---client model updates---be elements of a finite group, while the inputs are real-valued vectors represented with floating-point types. 
This requirement is usually addressed by converting client updates to fixed-point integers and operating in a finite domain (modulo~$2^p$) where $p$ is typically set in prior literature to 32 bits. The choice of \SecAgg bit-width~$p$ must balance communication costs with the accuracy loss due to rounding and overflows.

\para{Minimal Complexity.}
\looseness=-1
TEE-based protocols offer greater flexibility in how individual client updates can be processed; however, the code executed inside TEE is part of the trusted computing base (TCB) for all clients. In particular, it means that this code must be stable, auditable, defects- and side-channel-free, which severely limits its complexity. Hence, in practice, we prefer compression techniques that are either oblivious to \SecAgg's implementation or require minimal changes to the TCB.

\subsection{Compression Methods}
\label{subsec:comp_methods}
In this subsection, we consider a matrix $W \in \mathbb{R}^{\cin\times \cout}$ representing the weights of a linear layer to discuss three major compression methods with distinct compression/accuracy tradeoffs and identify the challenges \SecAgg faces to be readily amenable to these popular quantization algorithms.

\subsubsection{Scalar Quantization}
\label{subsec:sq}

\looseness=-1 Uniform scalar quantization maps floating-point weight $w$ to $2^b$ evenly spaced bins, where $b$ is the number of bits. Given a floating-point scale $s > 0$ and an integer shift parameter $z$ called the zero-point, we map any floating-point parameter $w$ to its nearest bin indexed by $\{0,\dots, 2^b-1\}$:
\[ w \mapsto \clamp(\round(w /s) + z, [0, 2^b - 1] ).\]
The tuple $(s, z)$ is often referred to as the quantization parameters (\texttt{qparams}).
With $b=8$, we recover the popular \texttt{int8} quantization scheme \citep{jacob2017quantization}, while setting $b = 1$ yields the extreme case of binarization \citep{courbariaux2015binaryconnect}. 
The quantization parameters $s$ and $z$ are usually calibrated after training a model with floating-point weights using the minimum and maximum values of each layer. 
The compressed representation of weights $W$ consists of the \texttt{qparams} and the integer representation matrix $W_q$ where each entry is stored in~$b$~bits. 
Decompressing any integer entry $w_q$ of~$W_q$ back to floating point is performed by applying  the (linear) operator $w_q \mapsto s\times(w_q - z)$.

\para{Challenge.} 
The discrete domain of quantized values and the finite group required by \SecAgg are not natively compatible because of the overflows that may occur at aggregation time. For instance, consider the extreme case of binary quantization, where each value is replaced by a bit. 
We can represent these bits in \SecAgg with $p=1$, but the aggregation will inevitably result in overflows.

\subsubsection{Pruning}
\label{subsec:rp}

Pruning is a class of methods that remove parts of a model such as connections or neurons according to some pruning criterion, such as weight magnitude~(\cite{lecun1990optimal,hassabi1992second}; see \cite{Blalock20} for a survey). \cite{konen2016federated} demonstrate client update compression with random sparsity for federated learning. Motivated by previous work and the fact that random masks do not leak information about the data on client devices, we will leverage random pruning of client updates in the remainder of this paper. 
A standard method to store a sparse matrix is the coordinate list (COO) format\footnote{See the  \href{https://pytorch.org/docs/stable/sparse.html}{torch.sparse documentation}.}, where only the non-zero entries are stored (in floating point or lower precision), along with their integer coordinates in the matrix. 
This format is compact, but only for a large enough compression ratio, as we store additional values for each non-zero entry.
Decompression is performed by re-instantiating the uncompressed matrix with both sparse and non-sparse entries.

\para{Challenge.}
Client update sparsity is an effective compression approach as investigated in previous work. However, the underlying assumption is that clients have different masks, either due to their seeds or dependency on client update parameters (\eg weight magnitudes). This is a challenge for \SecAgg as aggregation assumes a dense compressed tensor, which is not possible to construct when the coordinates of non-zero entries are not the same for all clients. 

\subsubsection{Product Quantization}
\label{subsec:pq}

Product quantization (PQ) is a compression technique developed for nearest-neighbor search \citep{jegou2011product} that can be applied for model compression \citep{stock2019bit}. 
Here, we show how we can re-formulate PQ to represent model updates. 
We focus on linear layers and refer the reader to~\cite{stock2019bit} for adaptation to convolutions.
Let the \emph{block size} be $d$ (say, 8), the number of \emph{codewords} be $k$ (say, 256) and assume that the number of input channels, $\cin$, is a multiple of $d$. 
To compress $W$ with PQ, we evenly split its columns into subvectors or blocks of size $d \times 1$ and learn a \emph{codebook} via $k$-means to select the $k$ codewords used to represent the $\cin\times\cout/d$ blocks of $W$. PQ with block size $d=1$ amounts to non-uniform scalar quantization with $\log_2 k$ bits per weight.

The PQ-compressed matrix $W$ is represented with the tuple $(C, A)$, where $C$ is the codebook of size $k \times d$ and $A$ gives the assignments of size $\cin \times\cout / d$.
Assignments are integers in $[0, k-1]$ and denote which codebook a subvector was assigned~to. 
To decompress the matrix (up to reshaping), we index the codebook with the assignments, written in PyTorch-like notation as
\begin{equation*}
    \widehat {W} = C[A].
\end{equation*}

\para{Challenge.}
There are several obstacles to making PQ compatible with \SecAgg.  
First, each client may have a different codebook, and direct access to these codebooks is needed to decode each client's message.  
Even if all clients share a (public) codebook, the operation to take assignments to produce an (aggregated) update is not linear, and so cannot be directly wrapped inside \SecAgg. 

\section{Method}
\label{sec:method}

In this section, we propose solutions to the challenges identified in Section~\ref{subsec:comp_methods} to reconcile security (\SecAgg) and communication efficiency (compression mechanisms).
Our approach is to modify compression techniques to share some hyperparameters globally across all clients so that aggregation can be done by uniformly combining each client's response, while still ensuring that there is scope to achieve accurate compressed representations.  

\subsection{Secure Aggregation and Compression}
\label{subsec:secagg_comp}
We propose to compress the uplink model updates through a compression operator $q$, whose parameters are round-dependent but the same for all clients participating in the same round. 
Then, we will add a random mask $m_i$ to each quantized client update $q(g_i)$ in the compressed domain, thus effectively reducing uplink bandwidth while ensuring that $h_i = q(g_i) + m_i$ is statistically indistinguishable from any other representable value in the finite group (see Section~\ref{subsec:secagg}). 
In this setting, \SecAgg allows the server to recover the aggregate of the client model updates in the compressed domain: $\sum_i q(g_i)$.
If the decompression operator $d$ is linear, the server is able to recover the aggregate in the non-compressed domain, up to quantization error, as illustrated in Figure~\ref{fig:secagg_summary}:
\begin{equation*}
    d\left(\sum_i h_i - \sum_i m_i\right) = d\left(\sum_i q(g_i)\right) =\sum_i d(q(g_i)) \approx \sum_i g_i.
\end{equation*}
The server periodically updates the quantization and decompression operator parameters, either from the aggregated model update, which is deemed public, or by emulating a client update on some similarly distributed public data. Once these parameters are updated, the server broadcasts them to the clients for the next round. This adds overhead to the downlink communication payload, however, this is negligible compared to the downlink model size. For instance, for scalar quantization, $q$ is entirely characterized by one \texttt{fp32} scale and one \texttt{int32} zero-point per layer, the latter of which is unnecessary in the case of a symmetric quantization scheme. Finally, this approach is compatible with both synchronous FL methods such as FedAvg~\citep{mcmahan2016communicationefficient} and asynchronous methods such as FedBuff~\citep{nguyen2021federated} as long as \SecAgg maintains the mapping between the successive versions of quantization parameters and the corresponding client updates.

\subsection{Application}
Next, we show how we adapt scalar quantization and random pruning with no changes required to \SecAgg. We illustrate our point with TEE-based \SecAgg while these adapted uplink compression mechanisms are agnostic of the \SecAgg mechanism. Finally, we show how to obtain extreme uplink compression by proposing a variant of \SecAgg, which we call \SecInd. This variant supports product quantization and is provably secure.

\subsubsection{Scalar Quantization and Secure Aggregation}
\label{subsubsec:sq_sa}

As detailed in Section~\ref{subsec:sq}, a model update matrix $g_i$ compressed with scalar quantization is given by an integer representation in the range $[0, 2^{b}-1]$ and by the quantization parameters \emph{scale} ($s$) and \emph{zero-point} ($z$). A sufficient condition for the decompression operator to be linear is to broadcast common quantization parameters per layer for each client. Denote $q(g_i)$ as the integer representation of quantized client model update $g_i$ corresponding to a particular layer for client $1\leq i \leq N$.
Set the scale of the decompression operator to $s$ and its zero-point to $z/N$. Then, the server is able to decompress as follows (recall that the decompression operator is defined in Section~\ref{subsec:sq}):
\begin{equation*}
    d\left(\sum_i q(g_i)\right) = s\sum_i  q(g_i) -  \frac{z}{N}  = \sum_i \left( s(q(g_i)) - z \right) \approx \sum_i g_i
\end{equation*}
Recall that all operations are performed in a finite group. Therefore, to avoid overflows at aggregation time, we quantize with a bit-width $b$ but take \SecAgg bit-width $p > b$, thus creating a margin for potential overflows (see Section~\ref{subsec:ablations}).
%
This approach is related to the fixed-point aggregation described in \citep{bonavitz2019federated,huba2021papaya}, but we calibrate the quantization parameters and perform the calibration per layer and periodically, unlike the related approaches.

\para{Privacy and Bandwidth.} Scales and zero points are determined from public data on the server. Downlink overhead is negligible: the server broadcasts the per-layer quantization parameters. The upload bandwidth is $p$ bits per weight, where $p$ is the \SecAgg finite group size (Section~\ref{subsec:secagg}).  
The maximum ``price of security'' is thus the overflow buffer of $p-b = \lceil\log_2 N\rceil$ bits per weight. 

\subsubsection{Pruning and Secure Aggregation}

To enable linear decompression with random pruning, all clients will share a common pruning mask for each round. 
This can be communicated compactly before each round as a seed for a pseudo-random function. 
This pruning mask seed is different from the \SecAgg mask seed introduced in Section~\ref{subsec:secagg} and has a distinct role.
Each client uses the pruning seed to reconstruct a pruning mask, prunes their model update $g_i$, and only needs to encrypt and transmit the unpruned parameters. 
The trade-off here is that some parameters are completely unobserved in a given round, as opposed to traditional pruning. 
\SecAgg operates as usual and the server receives the sum of the tensor of unpruned parameters computed by participating clients in the round, which it can expand using the mask seed.  
We denote the pruning operator as $\phi$ applied to the original model update $g_i$, and the decompression operator as $d$ applied to a compressed tensor $\phi(g_i)$. Decompression is an expansion operation equivalent to multiplication with a sparse permutation matrix $P_i$ whose entries are dependent on the $i$'th client's mask seed. 
Crucially, when all clients share the same mask seed within each round, we have $P_i = P$ for all $i$ and linearity of decompression is maintained:
\begin{equation*}
    d \left(\sum_i \phi(g_i) \right) = P \left( \sum_i \phi(g_i) \right) = \sum_i P_i\phi(g_i) = \sum_i d(\phi(g_i)) \approx \sum_i g_i.
\end{equation*}
\para{Privacy and Bandwidth.} Since the mask is random, no information leaks from the pruning mask. The downlink overhead (the server broadcasts one integer mask seed) is negligible. The upload bandwidth is simply the size of the sparse client model updates.

\subsubsection{Product Quantization and Secure Indexing}

\begin{algorithm}[t]
\caption{Secure Indexing (\SecInd)}
\label{alg:sec_indexing}
\begin{algorithmic}[1]


\Procedure{SecureIndexing}{C}      \Comment{This happens inside the TEE}
    \State Receive common codebook $C$ from server \Comment{$C$ is periodically updated by the server}
    \State Initialize histograms $H_{m,n}$ to $0$ \Comment{Each histogram for block $(m, n)$ has size $k$}
    \For{each client $i$}
    \State Receive and decrypt assignment matrix $A^i$ 
        \For{each block index $(m, n)$}
            \State $r \leftarrow A^i_{m, n}$ \Comment{Recover assignment of client $i$ for block $(m, m)$}
            \State $H_{m, n}[r] \leftarrow H_{m, n}[r] + 1$ \Comment{Update global count for codeword index $r$}
        \EndFor
    \EndFor
    \State Send back histograms $H_{m, n}$ to the server 
\EndProcedure
\end{algorithmic}
\end{algorithm}


We next describe the Secure Indexing (\SecInd) primitive, and discuss how to instantiate it. 
Recall that with PQ, each layer has its own codebook $C$ as explained in Section~\ref{sec:method}. 
Let us fix one particular layer compressed with codebook $C$, containing $k$ codewords. 
We assume that $C$ is common to all clients participating in the round. 
Consider the assignment matrix of a given layer $(A^i)_{m,n}$ for client~$i$.
From these, we seek to build the \emph{assignment histograms} $H_{m,n} \in \mathbb R^k$ that satisfy
\begin{equation*}
H_{m,n}[r] = \sum_i \mathbf 1\left(A^i_{m,n} = r\right),
\end{equation*}
where the indicator function $\mathbf 1$ satisfies $\mathbf 1\left(A^i_{m,n} = r\right) = 1$ if $A^i_{m,n} = r$ and $0$ otherwise.
A \emph{Secure Indexing} primitive will produce~$H_{m,n}$ while ensuring that no other information about client assignments or partial aggregations is revealed. 
The server receives assignment histograms from \SecInd and is able to recover the aggregated update for each block indexed by $(m, n)$ as 
\begin{equation*}
        \sum_r  H_{m,n}[r] \cdot C[r].
\end{equation*}
We describe how \SecInd can be implemented with a TEE
in Algorithm~\ref{alg:sec_indexing}.
Each client encrypts the assignment matrix, for instance with additive masking as described in Section~\ref{subsec:secagg}, and sends it to the TEE via the server. 
Hence, the server does not have access to the plaintexts client-specific assignments. 
TEE decrypts each assignment matrix and for each block indexed by $(m, n)$ produces the assignment histogram. 
Compared to \SecAgg, where the TEE receives an encrypted seed per client (a few bytes per client) and sends back the sum of the masks $m_i$ (same size as the considered model), \SecInd receives the (masked) assignment matrices and sends back histograms for each round. \SecInd implementation feasibility is briefly discussed in Appendix~\ref{appendix:secind}.


\para{Privacy and Bandwidth.} 
Codebooks are computed from public data while individual assignments are never revealed to the server. 
The downlink overhead of sending the codebooks is negligible as demonstrated in Section~\ref{sec:experiments}. 
The upload bandwidth in the TEE implementation is the assignment size, represented in $k$ bits (the number of codewords). 
For instance, with a block size $d=8$ and $k=32$ codewords, assignment storage costs are 5 bits per 8 weights, which converts to 0.625 bits per weight.
The tradeoff compared to non-secure PQ is the restriction to a global codebook for all clients (instead of one tailored to each client), and the need to instantiate \SecInd instead of \SecAgg.

\section{Experiments}
\label{sec:experiments}

\begin{figure*}[t]
    \centering
    \includegraphics[width=\textwidth]{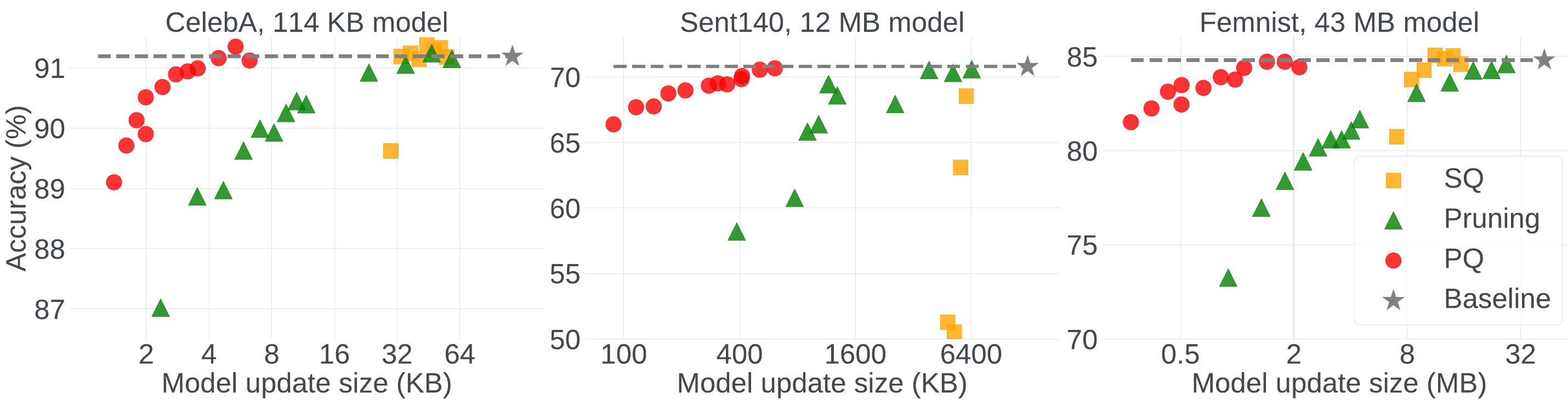}
    \vspace{-5mm}
    \caption{\label{fig:results_summary}
    We adapt scalar quantization (SQ) and pruning to the \SecAgg protocol to enable efficient and secure uplink communications. We also present results for product quantization (PQ) under the proposed novel \SecInd protocol. \emph{The $x$ axis is log-scale}. Baseline refers to \SecAgg FL run without any uplink compression, displayed as a horizontal line for easier comparison. Model size is indicated in the plot titles. Uncompressed client updates are as large as the models when $p=32$ (see Section~\ref{subsec:secagg}, represented as stars). We refer to the Appendix~\ref{appendix:table} for the matching tables where we additionally report the standard deviation of each data point. }
\end{figure*}

In this section, we numerically evaluate the performance of the proposed approaches when adapted to \SecAgg protocols. 
We study the relationship between uplink compression and model accuracy for the LEAF benchmark tasks. In addition, for scalar and product quantization we also analyze the impact of refresh rate for compression parameters on overall model performance.

\subsection{Experimental Setup}
\label{subsec:setup}
We closely follow the setup of~\cite{nguyen2021federated} and use the {FLSim library}
for our experiments\footnote{Code available at \texttt{github.com/facebookresearch/SecureFLCompression}.}. All experiments are run on a single V100 GPU 16 GB (except for Sent140 where we use one V100 32 GB) and typically take a few hours to run. More experiment details can be found in Appendix \ref{appendix:exp_details}.

\para{Tasks.} We run experiments on three datasets from the LEAF benchmark~\citep{caldas2018leaf}: CelebA~\citep{liu2015faceattributes}, Sent140~\citep{Go_Bhayani_Huang_2009} and FEMNIST~\citep{lecun2010mnist}. For CelebA, we train the same convolutional classifier as ~\citet{nguyen2021federated} with BatchNorm layers replaced by GroupNorm layers and 9,343 clients. For Sent140, we train an LSTM classifier for binary sentiment analysis with $59,400$ clients. Finally, for FEMNIST, we train a GroupNorm version of the ResNet18~\citep{he2015deep} for digit classification with 3,550 clients. For all compression methods, we do not compress biases and norm layers for their small overhead.

\para{Baselines.}  We focus here on the (synchronous) FedAvg approach although, as explained in Section~\ref{sec:method}, the proposed compression methods can be readily adapted to asynchronous FL aggregation protocols. 
We report the average and standard deviation of accuracy over three independent runs for all tasks at different uplink byte sizes corresponding to various configurations of the compression operator.


\para{Implementation Details.} We refer the reader to Appendix~\ref{appendix:exp_details} for all experiment details. 
The downlink overhead of sending the per-layer codebooks for product quantization is negligible as shown in Appendix~\ref{appendix:codebook_size}. Finally, the convergence time in terms of rounds is similar for PQ runs and the non-compressed baseline, as illustrated in Appendix~\ref{appendix:convergence}. Note that outside a simulated environment, the wall-clock time convergence for PQ runs would be \emph{lower} than the baseline since uplink communication would be more efficient, hence faster.

\subsection{Results and Comparison with Prior Work}

Results for efficient and secure uplink communications are displayed in Figure~\ref{fig:results_summary}. 
We observe that PQ yields a consistently better trade-off curve between model update size and accuracy. For instance, on CelebA, PQ achieves $\times 30$ compression with respect to the non-compressed baseline at iso-accuracy. The iso-accuracy compression rate is $\times 32$ on Sent140 and $\times 40$ on FEMNIST (see Appendix~\ref{appendix:table} for detailed tables). Scalar quantization accuracy degrades significantly for larger compression rates due to the overflows at aggregation as detailed in Appendix~\ref{appendix:overflow}. Pruning gives intermediate tradeoffs between scalar quantization and product quantization.

The line of work that develops FL compression techniques mainly includes FetchSGD~\citep{rothchild2020fetchsgd} as detailed in Section~\ref{sec:related}, although the authors do not mention \SecAgg. Their results are not directly comparable to ours due to non-matching experimental setups (e.g., datasets and architectures). 
However, Figure 6 in the appendix of~\cite{rothchild2020fetchsgd} mentions upload compression rates at iso-accuracy that are weaker than those obtained here with product quantization.

\subsection{Ablation Studies}
\label{subsec:ablations}

\begin{figure*}[t]
    \label{mask-refresh}
    \centering
    \includegraphics[width=0.9\textwidth]{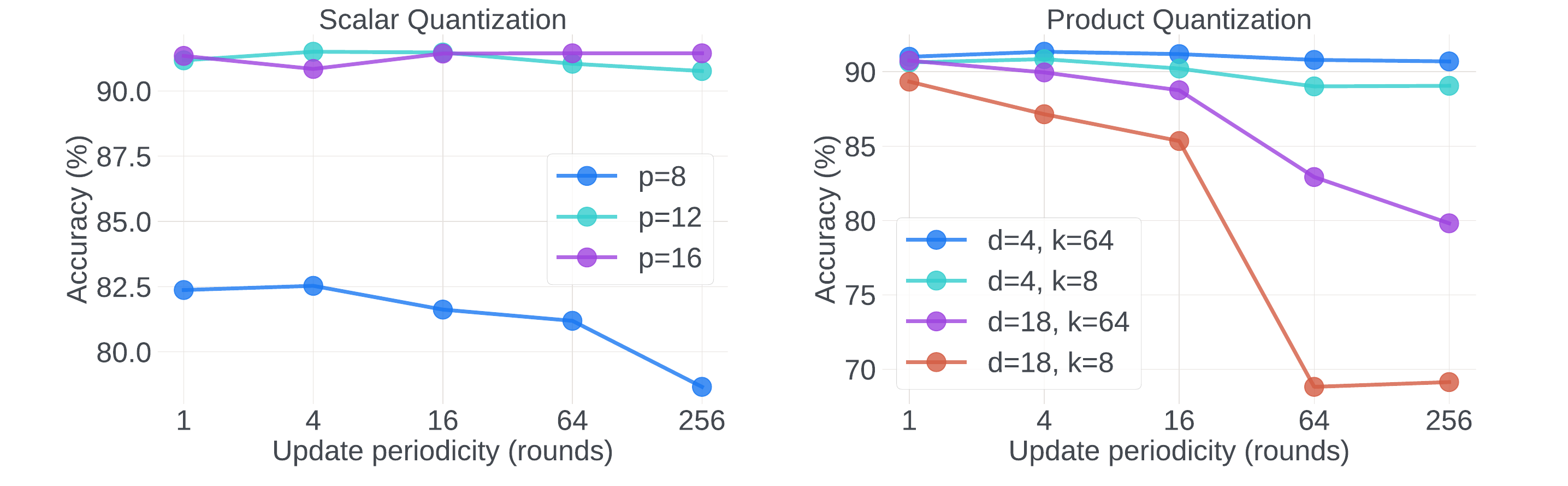}
    \caption{\label{fig:refresh}
    Impact of the refresh rate of the compression operator by the server on the CelebA dataset. \textbf{Left}: for scalar quantization (quantization parameters),  where we fix the quantization bit-width $b=8$ ($p$ denotes the \SecAgg bit-width). \textbf{Right}: for product quantization (codebooks), where $k$ denotes the number of codewords and $d$ the block size.}
\end{figure*}

We investigate the influence of the frequency of updates of the compression operator $q$ for scalar quantization and pruning, and study the influence of the \SecAgg bit-width $p$ on the number of overflows for scalar quantization.

\para{Update frequency of the compression operators.} In Figure~\ref{fig:refresh}, we show that for scalar quantization, the update periodicity only plays a role with low \SecAgg bit-width values $p$ compared to the quantization bit-width $b$. For product quantization, the update periodicity plays an important role for aggressive compression setups corresponding to large block sizes $d$ or to a smaller number of codewords $k$. For pruning, we measure the impact of masks that are refreshed periodically. 
We observe that if we refresh the compression operator more frequently, staleness is reduced, leading to accuracy improvements. We present our findings in Appendix~\ref{sparse_refresh}. 

\para{Overflows for scalar quantization.}
As discussed in Section~\ref{subsubsec:sq_sa}, we choose the \SecAgg bit-width~$p$ to be greater than quantization bit-width~$b$ in order to avoid aggregation overflows. While it suffices to set $p$ to be $\lceil\log_2 n_c\rceil$ more than $b$, where $n_c$ is the number of clients participating in the round, reducing $p$ is desirable to reduce uplink size. 
We study the impact of  $p$ on the percentage of parameters that suffer overflows and present our findings in Appendix~\ref{appendix:overflow}.

\section{Limitations and Future Work} 
\label{sec:discussion}

As mentioned in Section~\ref{sec:intro}, we need both \SecAgg and Differential Privacy~\citep{abadi2016deep} to realize the full promise of FL as a privacy-enhancing technology. While our primary focus is on enabling efficient uplink communication with \SecAgg protocols, we emphasize that the proposed approaches are compatible with DP. For instance, DP noise can be added natively by the TEE with our modified random pruning or scalar quantization approaches. 
For PQ  and \SecInd, it would require, however, to transfer the aggregation to TEE or to design a DP mechanism in the assignment space, since DP noise must be added by the TEE and not by the server.

A separate line of work tackles the challenge of combining communication efficiency and privacy. For instance, \cite{triastcyn2021dprec} develop a method that unifies compressed communication and DP (where integration with \SecAgg is left as an open problem), while \cite{chaudhuri2022privacyaware} design a privacy-aware scalar compression mechanism within the \emph{local} differential privacy model.

The broader social impact of this work is to enhance the privacy of individuals who may contribute to training ML models and simultaneously enable more individuals to participate in private training, who might otherwise have been excluded due to resource constraints. 




\section{Conclusion}
In this paper, we reconcile efficiency and security for uplink communication in Federated Learning. We propose to adapt existing compression mechanisms such as scalar quantization and pruning to the secure aggregation protocol by imposing a linearity constraint on the decompression operator. Our experiments demonstrate that we can adapt both quantization and pruning mechanisms to obtain a high degree of uplink compression with minimal degradation in performance and higher security guarantees. For achieving the highest rates of compression, we introduce \SecInd, a variant of \SecAgg well-suited for TEE-based implementation that supports product quantization wáhile maintaining a high security bar. 
We plan to extend our work to other federated learning scenarios, such as asynchronous FL, and further investigate the interaction of compression and privacy.



\bibliography{main}
\bibliographystyle{plainnat}
\section*{Checklist}


\begin{enumerate}
\item For all authors...
\begin{enumerate}
  \item Do the main claims made in the abstract and introduction accurately reflect the paper's contributions and scope?
    \answerYes{See the claim list in Section~\ref{sec:intro}}
  \item Did you describe the limitations of your work?
    \answerYes{See in particular Section~\ref{sec:discussion}}
  \item Did you discuss any potential negative societal impacts of your work?
    \answerYes{See in particular Section~\ref{sec:discussion}}
  \item Have you read the ethics review guidelines and ensured that your paper conforms to them?
    \answerYes{}
\end{enumerate}

\item If you are including theoretical results...
\begin{enumerate}
  \item Did you state the full set of assumptions of all theoretical results?
    \answerNA{}
        \item Did you include complete proofs of all theoretical results?
    \answerNA{}
\end{enumerate}

\item If you ran experiments...
\begin{enumerate}
  \item Did you include the code, data, and instructions needed to reproduce the main experimental results (either in the supplemental material or as a URL)?
    \answerYes{We provide detailed instructions in Section~\ref{sec:experiments} and in the Appendix}  \answerNo{We did not provide the code in the supplementary but will provide it when publishing the paper} 
  \item Did you specify all the training details (e.g., data splits, hyperparameters, how they were chosen)?
    \answerYes{See Section~\ref{sec:experiments} and in the Appendix}
        \item Did you report error bars (e.g., with respect to the random seed after running experiments multiple times)?
    \answerYes{See Section~\ref{sec:experiments} in the setup description and Appendix for the detailed error bars over 3 independent runs. Note that we do not report bars for ablation results.}
        \item Did you include the total amount of compute and the type of resources used (e.g., type of GPUs, internal cluster, or cloud provider)?
    \answerYes{See Section~\ref{sec:experiments} in the setup description.}
\end{enumerate}

\item If you are using existing assets (e.g., code, data, models) or curating/releasing new assets...
\begin{enumerate}
  \item If your work uses existing assets, did you cite the creators?
    \answerYes{See Section~\ref{sec:experiments}}
  \item Did you mention the license of the assets?
    \answerYes{See Appendix~\ref{appendix:exp_details}}
  \item Did you include any new assets either in the supplemental material or as a URL?
    \answerNo{}
  \item Did you discuss whether and how consent was obtained from people whose data you're using/curating?
    \answerNA{}
  \item Did you discuss whether the data you are using/curating contains personally identifiable information or offensive content?
    \answerNA{}
\end{enumerate}

\item If you used crowdsourcing or conducted research with human subjects...
\begin{enumerate}
  \item Did you include the full text of instructions given to participants and screenshots, if applicable?
    \answerNA{}
  \item Did you describe any potential participant risks, with links to Institutional Review Board (IRB) approvals, if applicable?
    \answerNA{}
  \item Did you include the estimated hourly wage paid to participants and the total amount spent on participant compensation?
    \answerNA{}
\end{enumerate}

\end{enumerate}

\appendix
\clearpage 
\section{Appendix}

\subsection{Experimental Details}
\label{appendix:exp_details}
In this section, we provide further details of the experimental setup described in Section~\ref{subsec:setup} and the hyper-parameters used for all the runs in Table~\ref{tab:hp}. For all the tasks, we use a mini-batch \SGD optimizer for local training at the client and \FedAvg optimizer for global model update on the server. The LEAF benchmark is released under the BSD 2-Clause License.

\paragraph{Baselines.} We run hyper-parameter sweeps to tune the client and server learning rates for the uncompressed baselines. Then, we keep the same hyper-parameters in all the runs involving uplink compression. We have observed that tuning the hyper-parameters for each compression factor does not provide significantly different results than using those for the uncompressed baselines, in addition to the high cost of model training involved. 

\paragraph{Compression details.}For scalar quantization, we use per-tensor quantization with MinMax observers. We use the symmetric quantization scheme over the integer range $[-2^{b-1}, 2^{b-1} - 1]$. For pruning, we compute the random mask separately for each tensor, ensuring all pruned layers have the same target sparsity in their individual updates. For product quantization, we explore various configurations by choosing the number of codewords per layer $k$ in $\{8, 16, 32, 64\}$ and the block size~$d$ in $\{4, 9, 18\}$. We automatically adapt the block size for each layer to be the largest allowed one that divides $C_{\text{in}}$ (in the fully connected case).

\begin{table}[!ht]
 \caption{Hyper-parameters used for all the experiments including baselines. $\eta$ is the learning rate.}
    \centering
    \begin{tabular}{l|ccccc}
    \toprule
    Dataset & Users per round & Client epochs & Max. server epochs & $\eta_\mathrm{\SGD}$ & $\eta_\mathrm{\FedAvg}$ \\
    \midrule
    CelebA & 100 & 1 & 30 & 0.90 & 0.08 \\
    Sent140 & 100 & 1 & 10 & 5.75 & 0.24 \\
    FEMNIST & 5 & 1 & 5 & 0.01 & 0.24 \\
    \bottomrule
    \end{tabular}
    \label{tab:hp}
\end{table}

\subsection{Experimental Results}
\label{appendix:results}

We provide various additional experimental results that are referred to in the main paper.
\subsubsection{Tables Corresponding to Figure~\ref{fig:results_summary}}
\label{appendix:table}
We provide the detailed results corresponding to Figure~\ref{fig:results_summary} along with standard deviation over 3 runs in Tables~\ref{tab:fig2_sq}, \ref{tab:fig2_prune}, and \ref{tab:fig2_pq}.

\subsubsection{Aggregation overflows with Scalar Quantization}
\label{appendix:overflow}
We discussed the challenge of aggregation overflows of quantized values with restricted \SecAgg finite group size in Section~\ref{subsec:sq} and noted in Section~\ref{subsubsec:sq_sa} that it suffices for \SecAgg bit-width~$p$ to be greater than quantization bit-width $b$ by at most $\lceil\log_2 N\rceil$, where $N$ is the number of clients participating in a given round. However, the overflow margin increases the client update size by~$p - b$ per weight. To optimize this further, we explore the impact of $p$ on aggregation overflows and accuracy, and present the results in Table~\ref{tab:celeba_overflows}.
As expected, we observe a decrease in percentage of weights that overflow during aggregation with the increase in the overflow margin size. However, while there is some benefit to non-zero overflow margin size, there is no strong correlation between the overflow margin size and accuracy, indicating the potential to achieve better utility even in the presence of overflows.

\begin{table}[!ht]
    \caption{Percentage of aggregation overflows (among all model parameters) for the CelebA dataset over various SQ configurations. $b$ is Quantization bit-width, $p$ is \SecAgg bit-width, $p-b$ is overflow margin size in bits.}
    \centering
    \begin{tabular}{ccccc}
    \toprule
    $b$ & $p$ & $p - b$ & Overflows (\% of parameters) & Accuracy \\
    \midrule
        4 & 4 & 0 & 3.71$\pm$1.53 & 49.33$\pm$2.03 \\
        4 & 5 & 1 & 1.43$\pm$0.55 & 50.44$\pm$1.77 \\
        4 & 6 & 2 & 0.68$\pm$0.43 & 49.67$\pm$1.56 \\
        4 & 7 & 3 & 0.17$\pm$0.12 & 51.58$\pm$0.66 \\
        4 & 8 & 4 & 0.06$\pm$0.00 & 87.30$\pm$0.36 \\
        4 & 9 & 5 & 0.06$\pm$0.00 & 89.19$\pm$0.20 \\
        4 & 10 & 6 & 0.06$\pm$0.00 & 88.52$\pm$0.07 \\
        4 & 11 & 7 & 0.05$\pm$0.00 & 87.68$\pm$1.24 \\
        \midrule 
        8 & 8 & 0 & 2.28$\pm$0.11 & 82.11$\pm$0.90 \\
        8 & 9 & 1 & 1.06$\pm$0.06 & 90.49$\pm$0.27 \\
        8 & 10 & 2 & 0.39$\pm$0.04 & 90.97$\pm$0.50 \\
        8 & 11 & 3 & 0.14$\pm$0.01 & 91.08$\pm$0.45 \\
        8 & 12 & 4 & 0.06$\pm$0.00 & 91.29$\pm$0.13 \\
        8 & 13 & 5 & 0.04$\pm$0.00 & 90.49$\pm$0.93 \\
        8 & 14 & 6 & 0.02$\pm$0.00 & 91.31$\pm$0.24 \\
        8 & 15 & 7 & 0.01$\pm$0.00 & 91.19$\pm$0.33 \\
        \bottomrule
    \end{tabular}
    \label{tab:celeba_overflows}
\end{table}

\subsubsection{Weighted aggregation and Scalar Quantization}
\label{appendix:w_agg}
Following the setup of \cite{nguyen2021federated}, we weight each client update by the number of samples the client trained on. Denoting the weight associated with the client $i$ with $\omega_i$ and following the same notations as in Section~\ref{subsec:secagg_comp}, weighted update is obtained as $h_i = (q(g_i) \times \omega_i) + m_i$. Since this is a synchronous FL setup, we do not set staleness factor. This weighted aggregation has no impact on pruning and product quantization, but can lead to overflows with scalar quantization. Therefore, we skip the weighting of quantized parameters of client updates and only weight non-quantized parameters (such as bias). For completion, we study with unweighted aggregation of client updates (including bias parameters) for scalar quantization experiments and present the result in Table~\ref{tab:sq_uw}. As expected, these results are similar to the ones with weighted aggregation.

\subsubsection{PQ Codebook Size is Negligible}
\label{appendix:codebook_size}

We demonstrate in Table~\ref{tab:codebook_size} that the overhead of sending codebooks (for all layers) is negligible compared to the model size. When the model is very small (CelebA model is 114 KB), reducing $k$ and $d$ makes the overhead negligible without hurting performance.

\begin{table*}[t]
     \caption{Cost of broadcasting codebooks (for downlink communications) is negligible compared to model sizes. Recall that $k$ denotes the number of codebooks and $d$ the block size.}
    \centering
    \vspace{3pt}
    \begin{tabular}{c|ccc}
    \toprule
    Dataset & Codebook size $k$ & Block size $d$ & Codebooks size (\% of model size) \\
    \midrule 
    \multirow{4}{*}{CelebA} & 8 & 4 & 0.6 KB (0.5\%) \\
    & 8 & 18 & 2.5 KB (2.2\%) \\
    & 64 & 4 & 4.2 KB (3.7\%) \\
    & 64 & 18 & 14.6 KB (12.8\%) \\
    \midrule
    \multirow{4}{*}{Sent140} & 8 & 4 & 0.9 KB (0.0\%) \\
    & 8 & 18 & 2.3 KB (0.0\%) \\
    & 64 & 4 & 5.4 KB (0.0\%) \\
    & 64 & 18 & 15.4 KB (0.1\%) \\
    \midrule
    \multirow{4}{*}{FEMNIST} & 8 & 4 & 2.6 KB (0.0\%) \\
    & 8 & 18 & 11.2 KB (0.0\%) \\
    & 64 & 4 & 20.8 KB (0.0\%) \\
    & 64 & 18 & 89.8 KB (0.2\%) \\
    \bottomrule
    \end{tabular}
    \label{tab:codebook_size}
\end{table*}

\subsubsection{Convergence Curves}
\label{appendix:convergence}
\begin{figure*}[t]
    \centering
    \includegraphics[width=0.6\textwidth]{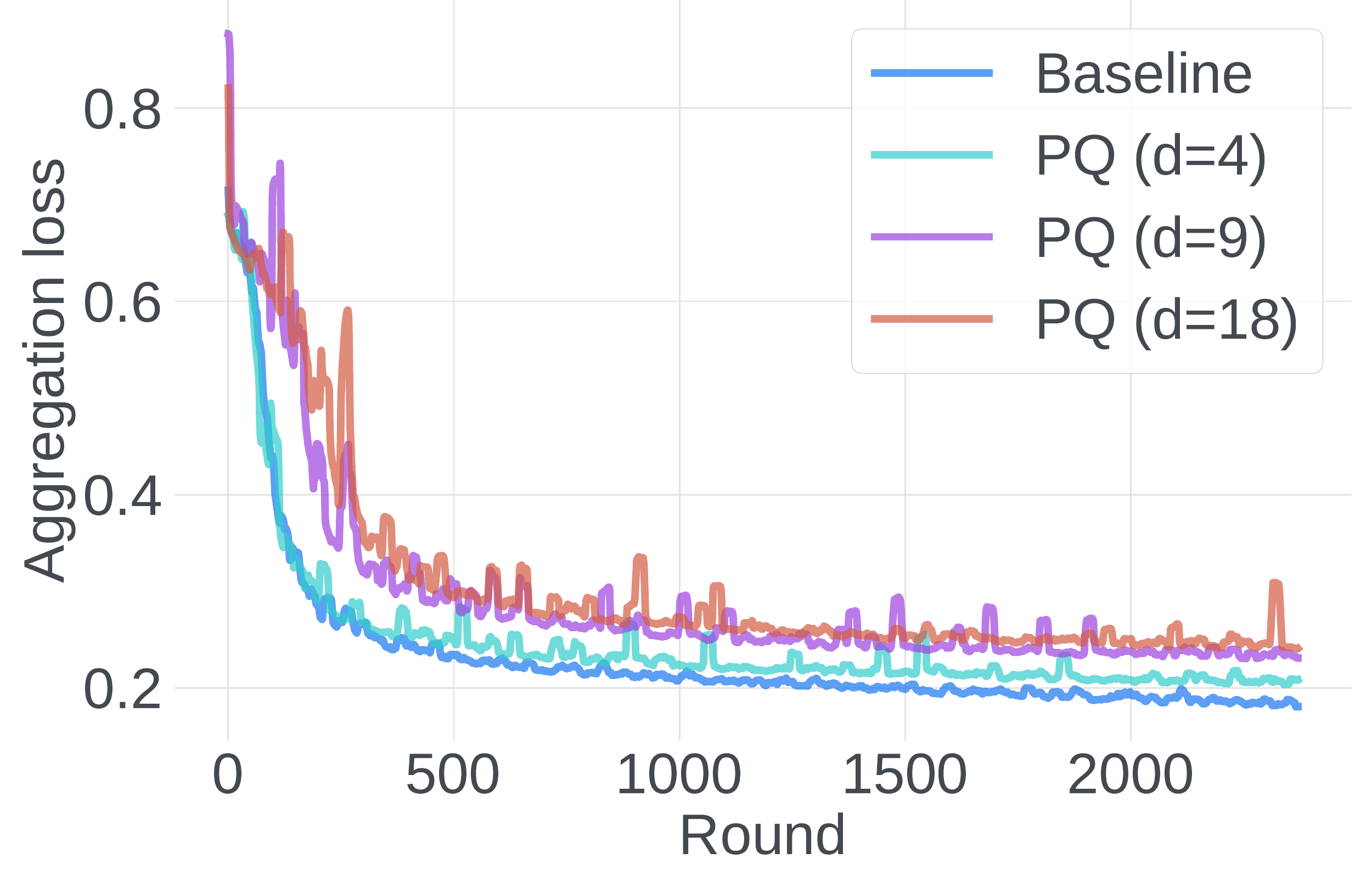}
    \caption{\label{fig:loss}
    Number of rounds to convergence is similar for PQ-compressed runs compared to the non-compressed baseline (on CelebA).
    Note that outside a simulated environment, the wall-clock time convergence for PQ runs would be lower than the baseline since uplink communications would be faster.}
\end{figure*}

We also provide convergence curves for PQ-compressed and baseline runs to demonstrate similar number of rounds needed to convergence in Figure~\ref{fig:loss}.

\subsubsection{Performance impact of sparsity mask refresh}
\label{sparse_refresh}

\begin{figure*}[t]
\label{fig-refresh}
    \centering
    \includegraphics[width=0.5\textwidth]{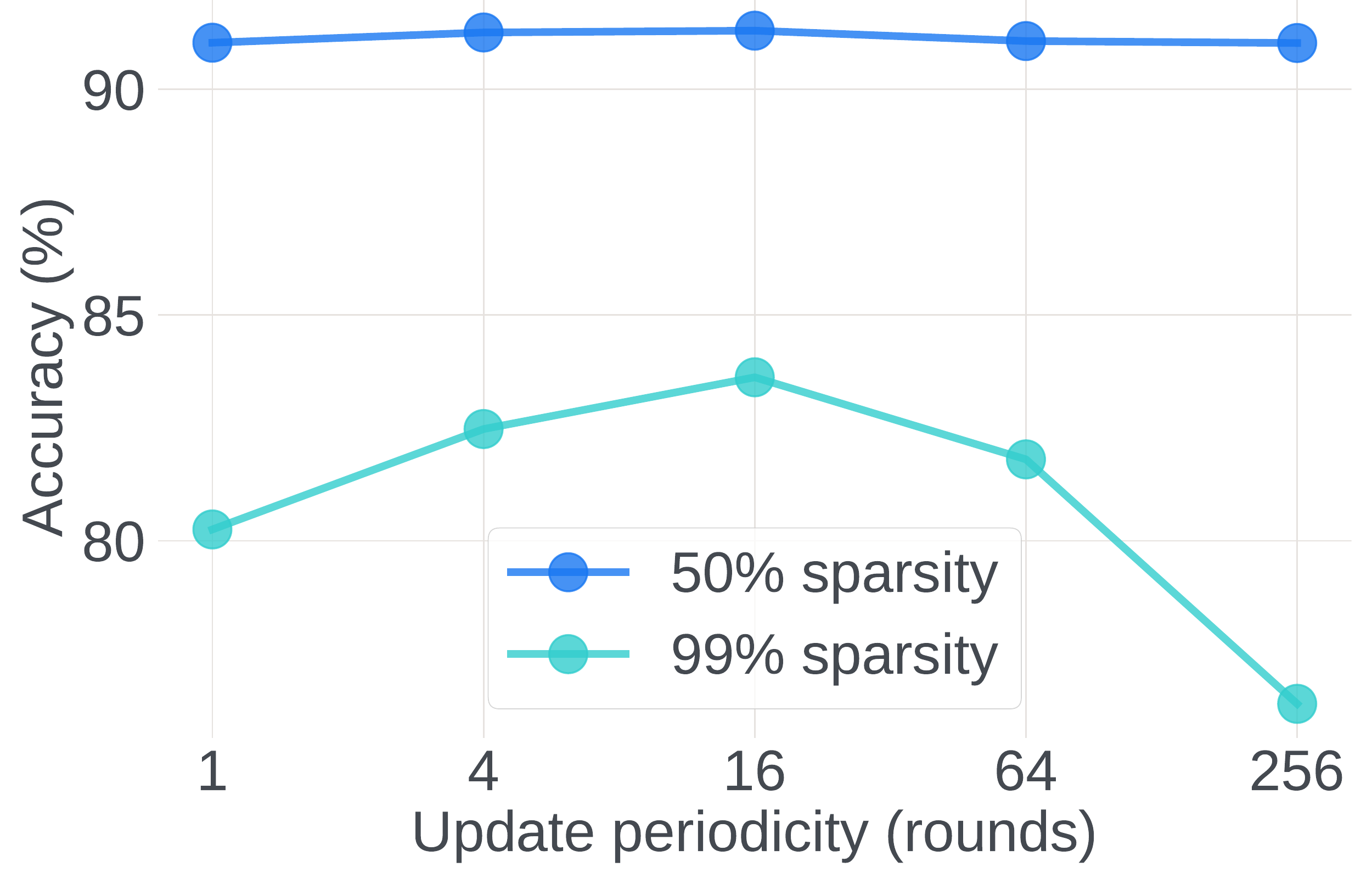}
    \caption{\label{fig:pruning}
    Impact of pruning mask refresh intervals on model performance for the CelebA dataset. Note that the effect of refreshing the pruning masks is more apparent at higher sparsity levels, and generalization performance decreases when masks are stale for longer during training. }
\end{figure*}

In addition to scalar and product quantization as described in Section~\ref{subsec:ablations}, we also conduct experiments with varying the interval for refreshing pruning masks. We consider two levels of sparsity, 50\% and 99\% and our experiments are on the CelebA dataset. We present our results in Figure~\ref{fig:pruning}. Overall we find that the model accuracy is robust to the update periodicity unless at very high sparsities, where accuracy decreases when mask refresh periodicity increases. This is important for future directions such as in asynchronous FL where clients have to maintain the same mask across successive global updates.

\subsection{\SecInd Implementations}
\label{appendix:secind}

\SecInd can be extended to other settings, such as multi-party computation (using two or more servers to operate on shares of the input), where each client can send evaluations of \emph{distributed point functions} to encode each assignment~\citep{boyle16}. 
These are represented compactly, but may require longer codewords to overcome the overheads. 
We leave the study of such software implementations of \SecInd to future work.

\begin{table}[!ht]
\caption{Results of client update compression with \SecAgg-compatible scalar quantization on LEAF datasets over three runs. We fix $p$ across runs as this defines the uplink size, but not $b$. We pick the run with the best accuracy and report the corresponding $b$.}
    \centering
    \begin{tabular}{l|rrrrr}
    \toprule
    Dataset & $b$ & $p$ & Uplink size (in KB) & Compression factor & Accuracy \\
    \midrule
        \multirow{15}{*}{\makecell{CelebA \\ Baseline: 91.2$\pm$0.2}} & 1 & 1 & 4.3 & 26.6 & 49.3$\pm$2.7 \\
        & 1,2 & 2 & 7.9 & 14.6 & 51.8$\pm$0.4 \\
        & 2,3 & 3 & 11.4 & 10.0 & 52.0$\pm$0.7 \\
        & 1,4 & 4 & 15.0 & 7.6 & 51.8$\pm$0.5 \\
        & 3,4 & 5 & 18.5 & 6.2 & 53.9$\pm$1.7 \\
        & 1,3,5 & 6 & 22.1 & 5.2 & 52.6$\pm$0.6 \\
        & 1,2 & 7 & 25.6 & 4.5 & 52.8$\pm$1.3 \\
        & 6 & 8 & 29.2 & 3.9 & 89.6$\pm$0.2 \\
        & 6,7 & 9 & 32.7 & 3.5 & 91.2$\pm$0.1 \\
        & 6,7 & 10 & 36.3 & 3.2 & 91.2$\pm$0.3 \\
        & 6,7,8 & 11 & 39.8 & 2.9 & 91.1$\pm$0.1 \\
        & 6,8 & 12 & 43.4 & 2.6 & 91.4$\pm$0.0 \\
        & 6,7 & 13 & 46.9 & 2.4 & 91.3$\pm$0.2 \\
        & 7,8 & 14 & 50.5 & 2.3 & 91.3$\pm$0.1 \\
        & 8 & 15 & 54.0 & 2.1 & 91.2$\pm$0.2 \\
        \midrule
        \multirow{15}{*}{\makecell{Sent140 \\ Baseline: 70.8$\pm$0.4}} & 1 & 1 &  399.3 & 31.5 & 46.2$\pm$0.0 \\
        & 1 & 2 &  792.3 & 15.9 & 53.8$\pm$0.0 \\
        & 2,3 & 3 &   1185.4 & 10.6 & 53.8$\pm$0.0 \\
        & 1,2 & 4 &  1578.4  & 8.0 & 53.8$\pm$0.0 \\
        & 2,3,5 & 5 &  1971.4  & 6.4 & 53.8$\pm$0.0 \\
        & 2,6 & 6 &  2364.5  & 5.3 & 53.8$\pm$0.0 \\
        & 1,7 & 7 &  2757.5  & 4.6 & 53.8$\pm$0.0 \\
        & 2,5 & 8 &  3150.6  & 4.0 & 53.8$\pm$0.0 \\
        & 5,6,7 & 9 &  3543.6  & 3.6 & 53.9$\pm$0.0 \\
        & 4,6 & 10 &  3936.6  & 3.2 & 53.8$\pm$0.1 \\
        & 6,8 & 11 &  4329.7  & 2.9 & 53.8$\pm$0.1 \\
        & 5,7 & 12 &  4722.7  & 2.7 & 51.3$\pm$4.4 \\
        & 6,8 & 13 & 5115.7 & 2.5 & 50.6$\pm$2.9 \\
        & 7 & 14 &  5508.8  & 2.3 & 63.1$\pm$3.1 \\
        & 8 & 15 &  5901.8  & 2.1 & 68.5$\pm$1.6 \\
        \midrule
        \multirow{11}{*}{\makecell{FEMNIST \\ Baseline: 84.8$\pm$0.7}} & 1 & 1 & 1404.0 & 31.2 & 2.2$\pm$3.0 \\
        & 1,2 & 2 & 2770.3 & 15.8 & 2.4$\pm$1.5 \\
        & 3 & 3 & 4136.5 & 10.6 & 12.2$\pm$3.9 \\
        & 4 & 4 & 5502.8 & 8.0 & 65.0$\pm$3.3 \\
        & 5 & 5 & 6869.0 & 6.4 & 80.7$\pm$0.4 \\
        & 6 & 6 & 8235.3 & 5.3 & 83.8$\pm$0.2 \\
        & 5,6,7 & 7 & 9601.5 & 4.6 & 84.3$\pm$0.4 \\
        & 6,7 & 8 & 10967.8 & 4.0 & 85.1$\pm$0.1 \\
        & 6,8 & 9 & 12334.1 & 3.6 & 84.9$\pm$0.2 \\
        & 7,8 & 10 & 13700.3 & 3.2 & 85.0$\pm$0.3 \\
        & 8 & 11 & 15066.6 & 2.9 & 84.6$\pm$0.3 \\
        \bottomrule
    \end{tabular}
    
    \label{tab:fig2_sq}
\end{table}

\begin{table}[!ht]
    \caption{Results of client update compression  with \SecAgg-compatible random mask pruning on LEAF datasets}
    \centering
    \begin{tabular}{l|rrrr}
    \toprule
    Dataset & Sparsity & Uplink size (in KB) & Compression factor & Accuracy \\
    \midrule
        \multirow{18}{*}{\makecell{CelebA \\ Baseline: 91.2$\pm$0.2}} & 0.1 & 103.0 & 1.1 & 91.2$\pm$0.2 \\
        & 0.2 & 91.6 & 1.2 & 91.3$\pm$0.0 \\
        & 0.3 & 80.3 & 1.4 & 91.1$\pm$0.2 \\
        & 0.4 & 68.9 & 1.7 & 91.1$\pm$0.2 \\
        & 0.5 & 57.5 & 2.0 & 91.1$\pm$0.1 \\
        & 0.6 & 46.1 & 2.5 & 91.2$\pm$0.1 \\
        & 0.7 & 34.8 & 3.3 & 91.1$\pm$0.1 \\
        & 0.8 & 23.4 & 4.9 & 90.9$\pm$0.2 \\
        & 0.9 & 12.0 & 9.5 & 90.4$\pm$0.2 \\
        & 0.91 & 10.9 & 10.5 & 90.4$\pm$0.1 \\
        & 0.92 & 9.8 & 11.7 & 90.3$\pm$0.2 \\
        & 0.93 & 8.6 & 13.3 & 89.9$\pm$0.2 \\
        & 0.94 & 7.5 & 15.3 & 90.0$\pm$0.3 \\
        & 0.95 & 6.3 & 18.1 & 89.6$\pm$0.1 \\
        & 0.96 & 5.2 & 22.0 & 89.0$\pm$0.4 \\
        & 0.97 & 4.1 & 28.2 & 88.9$\pm$0.1 \\
        & 0.98 & 2.9 & 39.1 & 87.0$\pm$0.5 \\
        & 0.99 & 1.8 & 64.1 & 83.8$\pm$0.2 \\
        \midrule
        \multirow{18}{*}{\makecell{Sent140 \\ Baseline: 70.8$\pm$0.4}} & 0.1 & 11325.7 & 1.1 & 70.5$\pm$0.3 \\
        & 0.2 & 10068.0 & 1.2 & 70.6$\pm$0.5 \\
        & 0.3 & 8810.3 & 1.4 & 70.6$\pm$0.3 \\
        & 0.4 & 7552.6 & 1.7 & 70.6$\pm$1.6 \\
        & 0.5 & 6294.9 & 2.0 & 70.6$\pm$0.8 \\
        & 0.6 & 5037.1 & 2.5 & 70.3$\pm$0.9 \\
        & 0.7 & 3779.4 & 3.3 & 70.5$\pm$0.7 \\
        & 0.8 & 2521.7 & 5.0 & 67.9$\pm$0.7 \\
        & 0.9 & 1264.0 & 10.0 & 68.6$\pm$1.8 \\
        & 0.91 & 1138.2 & 11.1 & 69.4$\pm$1.1 \\
        & 0.92 & 1012.4 & 12.4 & 66.4$\pm$3.1 \\
        & 0.93 & 886.7 & 14.2 & 65.8$\pm$2.5 \\
        & 0.94 & 760.9 & 16.5 & 60.8$\pm$0.0 \\
        & 0.95 & 635.1 & 19.8 & 51.6$\pm$5.7 \\
        & 0.96 & 509.4 & 24.7 & 52.7$\pm$3.9 \\
        & 0.97 & 383.6 & 32.8 & 58.2$\pm$7.8 \\
        & 0.98 & 257.8 & 48.8 & 60.3$\pm$5.9 \\
        & 0.99 & 132.0 & 95.3 & 49.3$\pm$4.1 \\
        \midrule
        \multirow{18}{*}{\makecell{FEMNIST \\ Baseline: 84.8$\pm$0.7}} & 0.1 & 39384.2 & 1.1 & 84.6$\pm$0.3 \\
        & 0.2 & 35010.3 & 1.2 & 84.7$\pm$0.3 \\
        & 0.3 & 30636.4 & 1.4 & 84.7$\pm$0.2 \\
        & 0.4 & 26262.5 & 1.7 & 84.6$\pm$0.2 \\
        & 0.5 & 21888.5 & 2.0 & 84.3$\pm$0.3 \\
        & 0.6 & 17514.7 & 2.5 & 84.2$\pm$0.2 \\
        & 0.7 & 13140.8 & 3.3 & 83.6$\pm$0.6 \\
        & 0.8 & 8766.9 & 5.0 & 83.1$\pm$0.2 \\
        & 0.9 & 4393.0 & 10.0 & 81.7$\pm$0.4 \\
        & 0.91 & 3955.6 & 11.1 & 81.1$\pm$0.4 \\
        & 0.92 & 3518.2 & 12.4 & 80.6$\pm$0.2 \\
        & 0.93 & 3080.8 & 14.2 & 80.6$\pm$0.2 \\
        & 0.94 & 2643.4 & 16.6 & 80.2$\pm$0.3 \\
        & 0.95 & 2206.0 & 19.8 & 79.4$\pm$0.1 \\
        & 0.96 & 1768.6 & 24.7 & 78.4$\pm$0.3 \\
        & 0.97 & 1331.2 & 32.9 & 77.0$\pm$0.5 \\
        & 0.98 & 893.9 & 49.0 & 73.3$\pm$0.4 \\
        & 0.99 & 456.5 & 95.9 & 65.6$\pm$0.1 \\
        \bottomrule
    \end{tabular}

    \label{tab:fig2_prune}
\end{table}

\begin{table}[!ht]
    \centering
        \caption{Results of client update compression with Product quantization and \SecInd on LEAF datasets}
    \begin{tabular}{l|rrrrr}
    \toprule
    Dataset & $k$ & $d$ & Uplink size (in KB) & Compression factor & Accuracy \\
    \midrule
        \multirow{12}{*}{\makecell{CelebA \\ Baseline: 91.2$\pm$0.2}} & 8 & 4 & 3.4 & 33.2 & 91.0$\pm$0.2 \\
        & 8 & 9 & 1.9 & 58.9 & 89.9$\pm$0.2 \\
        & 8 & 18 & 1.4 & 83.7 & 89.1$\pm$0.4 \\
        & 16 & 4 & 4.3 & 26.3 & 91.2$\pm$0.0 \\
        & 16 & 9 & 2.3 & 49.0 & 90.7$\pm$0.0 \\
        & 16 & 18 & 1.6 & 73.0 & 89.7$\pm$0.4 \\
        & 32 & 4 & 5.2 & 21.8 & 91.4$\pm$0.1 \\
        & 32 & 9 & 2.7 & 42.2 & 90.9$\pm$0.3 \\
        & 32 & 18 & 1.8 & 65.2 & 90.1$\pm$0.1 \\
        & 64 & 4 & 6.1 & 18.7 & 91.1$\pm$0.3 \\
        & 64 & 9 & 3.1 & 37.1 & 90.9$\pm$0.1 \\
        & 64 & 18 & 1.9 & 58.9 & 90.5$\pm$0.4 \\
        \midrule
        \multirow{12}{*}{\makecell{Sent140 \\ Baseline: 70.8$\pm$0.4}} & 8 & 4 & 301.0  &   41.8   & 69.5$\pm$1.5 \\
        & 8 & 9 & 204.2  &   61.6   & 69.0$\pm$0.8 \\
        & 8 & 18 & 86.3 &    145.8   & 66.4$\pm$0.3 \\
        & 16 & 4 & 399.3  &   31.5   & 69.8$\pm$1.1 \\
        & 16 & 9 & 270.2  &   46.6   & 69.3$\pm$0.1 \\
        & 16 & 18 &  113.0 &    111.3   & 67.7$\pm$0.6 \\
        & 32 & 4 & 497.5  &   25.3   & 70.6$\pm$0.2 \\
        & 32 & 9 & 336.2  &   37.4   & 69.4$\pm$0.4 \\
        & 32 & 18 & 139.7  &   90.1   & 67.7$\pm$2.3 \\
        & 64 & 4 & 595.8  &   21.1   & 70.7$\pm$0.3 \\
        & 64 & 9 & 402.2  &   31.3   & 70.1$\pm$1.0 \\
        & 64 & 18 & 166.4  &   75.6   & 68.7$\pm$0.7 \\
        \midrule
        \multirow{12}{*}{\makecell{FEMNIST \\ Baseline: 84.8$\pm$0.7}} & 8 & 4 &   1063.3   &  41.2       & 84.4$\pm$0.4 \\
        & 8 & 9 &  494.6   &  88.5       & 82.5$\pm$0.2 \\
        & 8 & 18 &   266.4  &   164.3       & 81.5$\pm$0.4 \\
        & 16 & 4 &   1405.1   &  31.1       & 84.7$\pm$0.2 \\
        & 16 & 9 &  646.8   &  67.6       & 83.3$\pm$0.1 \\
        & 16 & 18 &   342.6  &   127.7       & 82.2$\pm$0.5 \\
        & 32 & 4 &   1747.0   &  25.0       & 84.7$\pm$0.3 \\
        & 32 & 9 &  799.1   &  54.8       & 83.9$\pm$0.6 \\
        & 32 & 18 &   418.8  &   104.5       & 83.1$\pm$0.5 \\
        & 64 & 4 &   2088.9   &  20.9       & 84.4$\pm$0.2 \\
        & 64 & 9 &  951.4   &  46.0       & 83.8$\pm$0.8 \\
        & 64 & 18 &  495.1   &  88.4       & 83.5$\pm$0.7 \\
        \bottomrule
    \end{tabular}
    \label{tab:fig2_pq}
\end{table}

\begin{table}[!ht]
    \centering
        \caption{Results of scalar quantization on LEAF datasets with unweighted client update aggregation over three runs. We fix $p$ across runs as this defines the uplink size, but not $b$. We pick the run with the best accuracy and report the corresponding $b$.}
    \begin{tabular}{l|rrrrr}
    \toprule
    Dataset & $b$ & $p$ & Uplink size (in KB) & Compression factor & Accuracy \\
    \midrule
        \multirow{15}{*}{\makecell{CelebA \\ Baseline: 91.2$\pm$0.2}} & 1 & 1 & 4.3 & 26.5 & 50.3$\pm$1.92 \\
        & 1,2 & 2 & 7.9 & 14.6 & 50.9$\pm$0.54 \\
        & 1,2,3 & 3 & 11.4 & 10.0 & 52.0$\pm$0.46 \\
        & 2,3,4 & 4 & 15.0 & 7.6 & 51.8$\pm$0.17 \\
        & 1,3,4 & 5 & 18.5 & 6.2 & 52.6$\pm$1.23 \\
        & 3,4 & 6 & 22.1 & 5.2 & 52.8$\pm$1.21 \\
        & 2,3 & 7 & 25.6 & 4.5 & 51.9$\pm$0.05 \\
        & 6 & 8 & 29.2 & 3.9 & 90.2$\pm$0.19 \\
        & 6 & 9 & 32.7 & 3.5 & 90.8$\pm$0.04 \\
        & 6 & 10 & 36.3 & 3.2 & 91.2$\pm$0.08 \\
        & 6,7 & 11 & 39.8 & 2.9 & 91.2$\pm$0.20 \\
        & 6,8 & 12 & 43.4 & 2.6 & 91.2$\pm$0.11 \\
        & 6 & 13 & 46.9 & 2.4 & 91.4$\pm$0.13 \\
        & 7,8 & 14 & 50.5 & 2.3 & 91.3$\pm$0.15 \\
        & 8 & 15 & 54.0 & 2.1 & 91.3$\pm$0.22 \\
        \midrule
        \multirow{15}{*}{\makecell{Sent140 \\ Baseline: 70.8$\pm$0.4}} & 1 & 1 & 399.3 & 31.5 & 51.3$\pm$4.4 \\
        & 1,2 & 2 & 792.3 & 15.9 & 51.3$\pm$4.4 \\
        & 1,2 & 3 & 1185.4 & 10.6 & 53.8$\pm$0.0 \\
        & 1,2 & 4 & 1578.4 & 8.0 & 53.8$\pm$0.0 \\
        & 2,3 & 5 & 1971.4 & 6.4 & 53.8$\pm$0.0 \\
        & 1,5 & 6 & 2364.5 & 5.3 & 53.8$\pm$0.0 \\
        & 1,3,4 & 7 & 2757.5 & 4.6 & 53.8$\pm$0.0 \\
        & 2,4,7 & 8 & 3150.6 & 4.0 & 53.8$\pm$0.1 \\
        & 3,5 & 9 & 3543.6 & 3.6 & 53.8$\pm$0.0 \\
        & 4,5 & 10 & 3936.6 & 3.2 & 53.8$\pm$0.1 \\
        & 5,7 & 11 & 4329.7 & 2.9 & 52.0$\pm$3.2 \\
        & 6,7 & 12 & 4722.7 & 2.7 & 53.8$\pm$0.0 \\
        & 6 & 13 & 5115.7 & 2.5 & 60.3$\pm$2.9 \\
        & 7 & 14 & 5508.8 & 2.3 & 65.9$\pm$2.9 \\
        & 8 & 15 & 5901.8 & 2.1 & 67.7$\pm$0.9 \\
        \midrule
        \multirow{11}{*}{\makecell{FEMNIST \\ Baseline: 84.8$\pm$0.7}} & 1 & 1 & 1404.0 & 31.2 & 2.0$\pm$1.3 \\
        & 1,2 & 2 & 2770.3 & 15.8 & 4.8$\pm$2.8 \\
        & 1,2,3 & 3 & 4136.5 & 10.6 & 13.6$\pm$2.4 \\
        & 4 & 4 & 5502.8 & 8.0 & 66.3$\pm$3.5 \\
        & 5 & 5 & 6869.0 & 6.4 & 79.7$\pm$0.8 \\
        & 5 & 6 & 8235.3 & 5.3 & 84.0$\pm$0.9 \\
        & 6,7 & 7 & 9601.5 & 4.6 & 84.3$\pm$0.1 \\
        & 6,7,8 & 8 & 10967.8 & 4.0 & 84.8$\pm$0.6 \\
        & 7,8 & 9 & 12334.1 & 3.5 & 85.1$\pm$0.1 \\
        & 7,8 & 10 & 13700.3 & 3.2 & 85.0$\pm$0.4 \\
        & 8 & 11 & 15066.6 & 2.9 & 83.5$\pm$1.7 \\
        \bottomrule
    \end{tabular}
    \label{tab:sq_uw}
\end{table}



\end{document}